\documentclass[letterpaper]{article} 
\usepackage{aaai24}  
\usepackage{times}  
\usepackage{helvet}  
\usepackage{courier}  
\usepackage[hyphens]{url}  
\usepackage{graphicx} 
\usepackage{natbib}  
\usepackage{caption} 
\usepackage{algorithm}
\usepackage{algorithmic}

\usepackage{amsmath,amsfonts,bm}

\def\eqref#1{equation~\ref{#1}}

\def\1{\bm{1}}

\def\va{{\bm{a}}}

\def\vo{{\bm{o}}}

\def\vs{{\bm{s}}}

\def\vz{{\bm{z}}}

\DeclareMathAlphabet{\mathsfit}{\encodingdefault}{\sfdefault}{m}{sl}
\SetMathAlphabet{\mathsfit}{bold}{\encodingdefault}{\sfdefault}{bx}{n}

\usepackage{makecell}
\usepackage{multirow}
\usepackage{newfloat}
\usepackage{listings}
\DeclareCaptionStyle{ruled}{labelfont=normalfont,labelsep=colon,strut=off} 
\floatstyle{ruled}
\newfloat{listing}{tb}{lst}{}
\floatname{listing}{Listing}
\usepackage{color}
\nocopyright 

\title{Learning Robust Representations via Bidirectional Transition
for Visual Reinforcement Learning}
\author {
    Xiaobo Hu\textsuperscript{\rm 1},
    Youfang Lin\textsuperscript{\rm 1},
     Yue Liu\textsuperscript{\rm 1},
    Jinwen Wang\textsuperscript{\rm 1},
    Shuo Wang\textsuperscript{\rm 1},
   Hehe Fan\textsuperscript{\rm 2} and 
    Kai Lv\textsuperscript{\rm 1},
}
\affiliations {
    \textsuperscript{\rm 1}Beijing Jiaotong University
    \textsuperscript{\rm 2}Zhejiang University\\
    xiaobohu@bjtu.edu.cn, yflin@bjtu.edu.cn,liuyue@bjtu.edu.cn,jwwang@bjtu.edu.cn,shuo.wang@bjtu.edu.cn,hehefan@zju.edu.cn, lvkai@bjtu.edu.cn
}

\usepackage{bibentry}

\begin{document}

\maketitle

\begin{abstract}
Visual reinforcement learning has proven effective in solving control tasks with high-dimensional observations. 
However, extracting reliable and generalizable representations from vision-based observations remains a central challenge.
Inspired by the human thought process, when the representation extracted from the observation can predict the future and trace history, the representation is reliable and accurate in comprehending the environment.
Based on this concept, we introduce a \textbf{Bi}directional \textbf{T}ransition (BiT) model, which leverages the ability to bidirectionally predict environmental transitions both forward and backward to extract reliable representations.
Our model demonstrates competitive generalization performance and sample efficiency on two settings of the DeepMind Control suite.
Additionally, we utilize robotic manipulation and CARLA simulators to demonstrate the wide applicability of our method.
\end{abstract}

\section{Introduction}
Visual Reinforcement Learning (Visual RL) is an active research field and focuses on learning optimal control policies from high-dimensional image inputs.
Visual RL has shown remarkable success in various applications, including video games \cite{mnih2013playing}, robotic manipulation \cite{levine2016end}, and autonomous navigation \cite{10168913}.
However, generalizing the learned policies to novel environments with visual changes remains a significant challenge \cite{cobbe2019quantifying,zhang2018study}.

Recent studies have achieved significant progress in the generalization task by utilizing self-supervised methods~\cite{grill2020bootstrap,laskin2020curl}.
These methods design specific auxiliary tasks to imbue the representation with certain properties \cite{hansen2021generalization}, resulting in effective generalization.
Similarly, we focus on designing auxiliary tasks to extract representations for the generalization of visual reinforcement learning.

\begin{figure}
  \centering
  \includegraphics[width=1\linewidth]{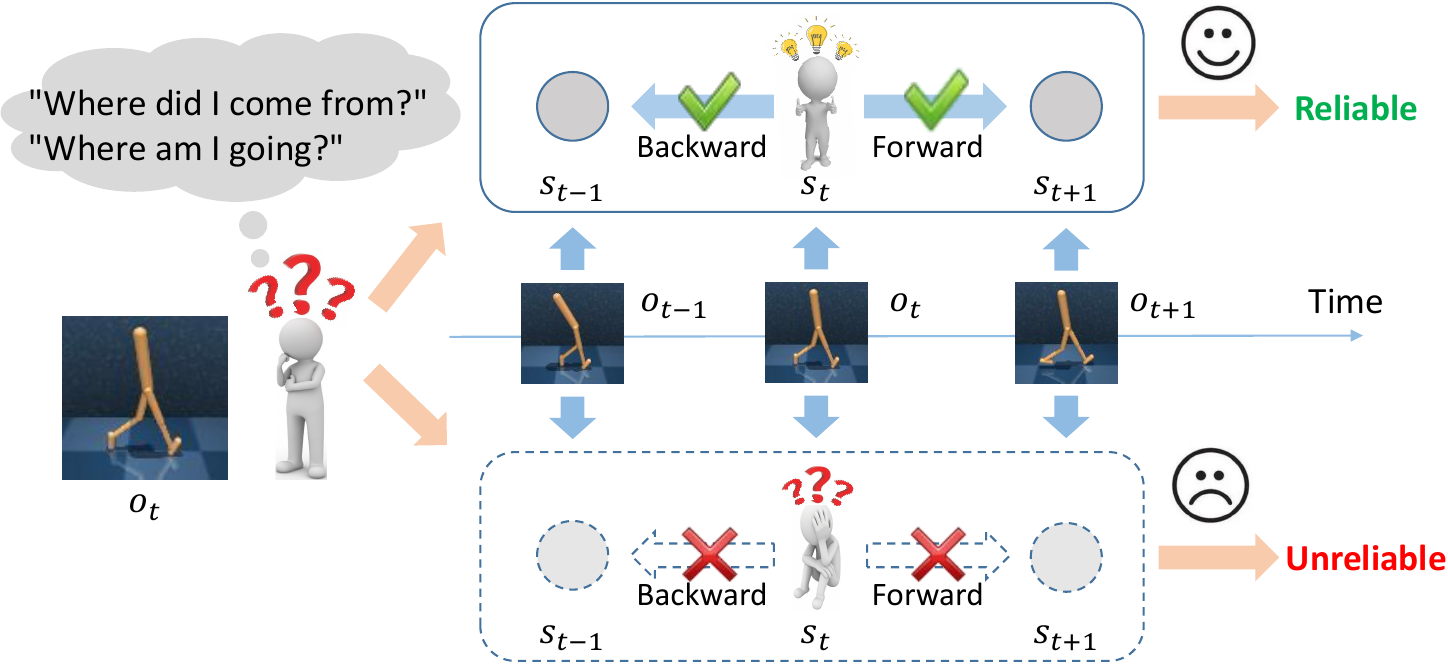}
  \caption{\textbf{The basic idea of our method.} When humans encounter a perplexing situation, they usually conduct both backward and forward environmental transition analysis. 
  A successful analysis indicates trustworthy comprehension of the current environment, while a failed analysis suggests unreliable environmental understanding.
  }
  \label{Intro}
\end{figure}

The auxiliary task in our method is to predict bidirectional environment transitions, aiding the agent in extracting more reliable representations.
The intuition behind our method is that representations with strong environmental transition properties can better comprehend the context information and provide more reliable support for decision-making.
In Figure \ref{Intro}, we draw parallels to the human thought process.
When humans encounter a perplexing decision-making situation and struggle to accurately assess their current state, they often ask themselves two crucial questions: ``Where did I come from?'' and ``Where am I going?''.
If they can successfully trace back history and predict the future based on the current situation, they will have a reliable comprehension of their current state, facilitating dependable decisions.
Conversely, if they cannot backtrack and deduce, it means a lack of proper comprehension of the situation, and making decisions based on this incorrect comprehension would also be unreliable.


Inspired by the above human thought, we propose a \textbf{Bi}directional \textbf{T}ransition (BiT) model to extract reliable representations from image observations for policy learning.
Specifically, we first design a feature extractor to generate representations of visual observations. 
Then, we propose a bidirectional transition learner that considers both forward and backward transition predictions. 
To avoid trivial solutions, our approach includes an action prediction module within the bidirectional transition learner.
Learning representation with BiT is independent of vision-based reinforcement learning methods and can be utilized with any policy learner.
In this paper, we implement the standard Soft Actor-Critic (SAC) \cite{haarnoja2018soft} as the policy learner.

We employ two distinct configurations of the DeepMind Control suite (DMControl) \cite{tassa2018deepmind,hansen2021generalization} to validate the generalization ability and sample efficiency of BiT. 
Additionally, we implement the robotic manipulation simulator \cite{jangir2022look} and driving simulator CARLA \cite{dosovitskiy2017carla} to illustrate  the wide applicability of our method.
Extensive experiments exhibit that BiT has superior generalization performance and sample efficiency compared to previous state-of-the-art methods.
In summary, our contributions are threefold: 
\begin{itemize}
    \item We introduce the concept of utilizing predictions of backward and forward transitions in visual RL for observation comprehension, aligning with human intuition.
    \item We propose a \textbf{Bi}directional \textbf{T}ransition (BiT) model that can assist the agent in obtaining stable and reliable representations from the image observations.
    \item Extensive experimental results show that our method achieves superior results with previous state-of-the-art methods in generalization and sample efficiency.
\end{itemize}

\section{Related Work}
\subsection{Data Augmentation in Visual RL}
Data Augmentation has been demonstrated to significantly improve the generalization of visual RL.
Randomized convolution \cite{DBLP:conf/iclr/LeeLSL20} is a simple technique to improve the generalization ability of agents introducing a randomized neural network to perturb input observations.
RAD \cite{laskin2020reinforcement} compares different data augmentation methods and evidences their benefits in visual RL.
DrQ \cite{yarats2020image} aggregates multiple image augmentations, averaging the value functions and their targets.
UCB-DrAC \cite{raileanu2021automatic} combines the previous method with UCB \cite{auer2002using} and introduces three approaches for automatically finding an effective augmentation.
SECANT \cite{pmlr-v139-fan21c} mentions the weak and strong augmentations and shows their different effects on the generalization task.
CNSN \cite{li2023normalization} proposes a normalization technique without relying on prior knowledge of the shift characteristics.
CLOP \cite{DBLP:conf/iclr/BertoinR22}  introduces a novel regularization approach involving channel-consistent local permutations to the feature maps. 
Data augmentation has demonstrated significant efficacy in enhancing generalization by leveraging prior knowledge as an inductive bias for the agent \cite{ma2022comprehensive}.

\subsection{Representation learning in Visual RL}
Numerous approaches have been dedicated to learning improved representations that can bridge the generalization gap in visual RL.
For example, DBC \cite{zhang2020learning} introduces a bisimulation metric \cite{ferns2011bisimulation} to learn robust representations. 
CURL \cite{laskin2020curl} implements contrastive learning \cite{chen2020simple} and data augmentations to learn representation more efficiently.
SVEA \cite{hansen2021stabilizing} identifies two problems rooted in high-variance Q-targets and proposes a technique for stabilizing this variance under data augmentations.
PAD \cite{DBLP:conf/iclr/HansenJSAAEPW21} proposes a method that utilizes a self-supervision task to allow the policy to continue training after deployment in a novel environment without relying on any rewards. 
SPD \cite{kim2022self} designs a self-predictive dynamics model as an auxiliary task to extract task-relevant features efficiently. 
However, SPD exclusively focuses on forward transition prediction, leading to the potential acquisition of unreliable state representations.
We not only account for predicting the forward transition but also consider the backward transition.
Our method can more effectively learn the environmental contextual information to obtain a reliable representation of observation.

\begin{figure*}
  \centering
  \includegraphics[width=1\linewidth]{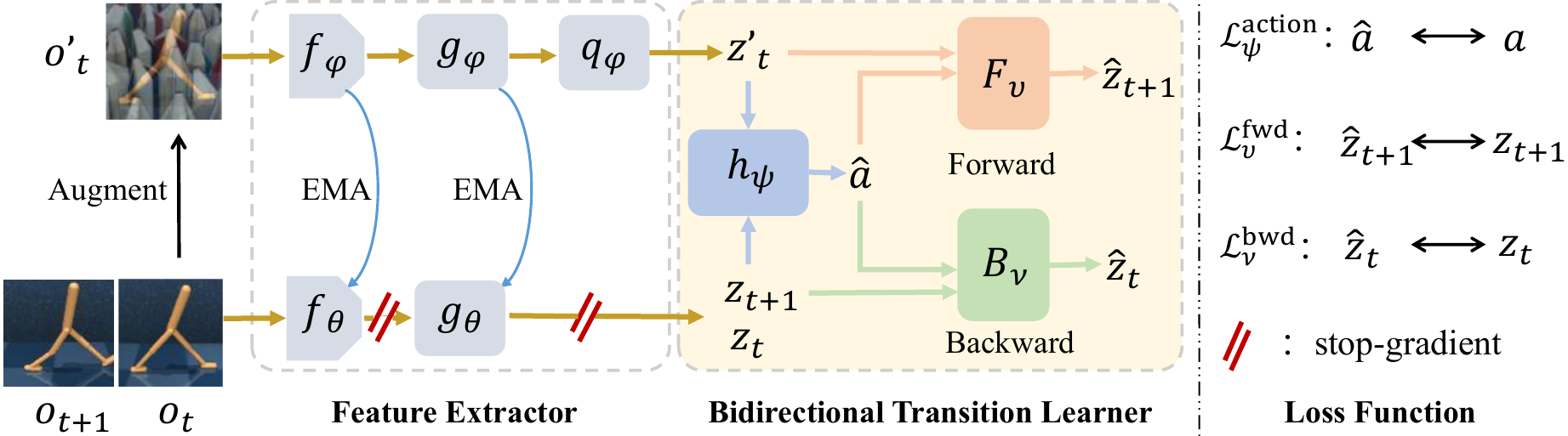}
  \caption{\textbf{Overview of our model.} 
  We implement a feature extractor with a parallel architecture to generate distinct encoded representations.
  For the bidirectional transition learner, there are three modules, \textit{i.e.,} the forward transition module $F_{\upsilon}$, the backward transition module $B_{\nu}$, and the action prediction module $h_{\psi}$.
  Our objective is to utilize the loss functions of the bidirectional transition prediction to learn the encoder $f_{\varphi}$ for generating reliable representations.
  }
  \label{OverFlow}
\end{figure*}

\subsection{Transition Models in RL}
Previous RL methods achieve superior sample efficiency by learning a transition model of the environment and planning their policy based on the model \cite{sutton2018reinforcement}. 
These algorithms benefit from the trained forward \cite{finn2017deep} or backward transition model \cite{wang2021offline} and also have better generalization. 
For example, MOPO \cite{yu2020mopo} proposes a modified method applying the existing transition models with rewards artificially penalized by the uncertainty of the transitions.
BMIL \cite{NEURIPS2022_7ce5da35} introduces a generative backward dynamics model and generates short imagined trajectories from states in the demonstrations.
Inspired by these transition model learning methods, we propose the bidirectional transition model learning method in high-dimensional vision-based reinforcement learning tasks.

\section{Preliminaries}
\subsection{Visual Reinforcement Learning}
We formulate the interaction between the agent and environment as a Partially Observable Markov Decision Process (POMDP) \cite{kaelbling1998planning} $\mathcal{M} =\left \langle \mathcal{S},\mathcal{O},\mathcal{A},\mathcal{P},r,\mathcal{\gamma } \right \rangle$, 
where $\mathcal{S}$ is the state space, 
$\mathcal{O}$ is the high-dimensional observation space \cite{mnih2013playing}, 
$\mathcal{A}$ is the action space,
$\mathcal{P}(\vs_{t+1}|\vs_{t}, \va_{t})$ is the state transition function,
$r:\mathcal{S}\times \mathcal{A}\mapsto  \mathbb{R} $ is the scalar reward function, 
and $\gamma \in \left [  0,1 \right ) $ is the discount factor.
RL aims to train an agent policy $\pi_{\theta } \left ( \cdot | s\right ) $ parameterized by $\theta$ and maximize the expected cumulative return $ J(\theta )=\mathbb{E}_{\va_{t}\sim \pi _{\theta }\left (\cdot |\vs_{t} \right ),\vs_{t}\sim\mathcal{P} }\left [  {\textstyle \sum_{t=0}^{T}\gamma ^{t}r\left (  \vs_{t},\va_{t}\right )  }  \right ] $, where $T$ is the horizon of the POMDP.
In the visual generalization setting, we expect the learned policy $\pi_{\theta }$ to be well generalized to novel environments with the same structure as the original POMDP but with different observation space $\tilde{\mathcal{O}}$ constructed from the same state space $\mathcal{S}$.

\subsection{Environment Transition Model}
The basic transition models in RL learn the environment transition, and others also learn the reward function simultaneously.
Categorized by the transfer direction, it can be divided into the forward transition model $p_{\upsilon }\left ( \vs_{t+1},r_{t}| \vs_{t},\va_{t}\right ) $ and the backward transition model $p_{\nu }\left (\vs_{t},r_{t}| \vs_{t+1},\va_{t}\right ) $ parameterized by $\upsilon $ and $\nu$, respectively.
The forward model $p_{\upsilon }$ represents the probability of the next state and corresponding reward given the current state and action.
Conversely, the backward model $p_{\nu}$ generates the probability for the current state and reward by taking the next state and action as input. 
The reward function $r\left ( \vs_{t},\va_{t} \right ) $ only depends on the current state $\vs_{t}$ and action $\va_{t}$. 
So the forward transition can be decomposed as $p_{\upsilon }( \vs_{t},r_{t}| \vs_{t+1},\va_{t} ) =p(\vs_{t+1}|\vs_{t},\va_{t})p(r_{t}|\vs_{t},\va_{t})$ and backward transition as well \cite{lyu2022double}.
The maximizing of the log-likelihood functions are as follows:
\begin{equation}
\mathcal{L}_{\upsilon }^{fwd}=\mathbb{E}_{(\vs_{t},\va_{t},r_{t},\vs_{t+1})\sim \mathcal{D}_{\mathrm{env}}}[-\log{ p_{\upsilon}{(\vs_{t+1},r_{t}|\vs_{t},\va_{t})}}],
\end{equation}
\begin{equation}
\mathcal{L}_{\nu }^{bwd}=\mathbb{E}_{(\vs_{t},\va_{t},r_{t},\vs_{t+1})\sim \mathcal{D}_{\mathrm{env}}}[-\log{p_{\nu}}{(\vs_{t},r_{t}|\vs_{t+1},\va_{t})} ],
\end{equation}
where $\mathcal{D}_{\mathrm{env}}$ is the sample dataset of the agent policy.

\section{Method}
We present a \textbf{Bi}directional \textbf{T}ransition (BiT) Model, depicted in Figure~\ref{OverFlow} and described in Algorithm~\ref{alg1}. 
Our approach ensures that the representation reliably comprehends the current visual observation by leveraging predictions from a bidirectional transition learner.
Notably, our method provides an auxiliary representation learning task, effectively assisting shared encoders' updates to offer dependable representations for reinforcement learning.
BiT is a generic model adaptable to various vision-based RL algorithms.

\subsection{Overview}
Our model aims to deliver a reliable and generalizable representation from the input observation for the policy learner.
Figure \ref{OverFlow} illustrates that BiT comprises a feature extractor and a bidirectional transition learner.
During training, we introduce data augmentation to perturb image observations. 
Subsequently, we leverage the feature extractor with a parallel architecture \cite{grill2020bootstrap} to generate distinct encoded representations. 
The bidirectional transition learner mainly consists of three parts: the forward transition module $F_{\upsilon}$, the backward transition module $B_{\nu}$, and an action prediction module $h_{\psi}$. 
For the policy learner, we implement the standard Soft Actor-Critic (SAC)~\cite{haarnoja2018soft} algorithm, sharing the encoder $f_{\varphi}$ and utilizing the unaugmented observations $\vo_t$ to learn the control policy.
BiT and the policy learner are alternating training, as shown in Algorithm~\ref{alg1}.
During testing, we only utilize the encoder $f_{\varphi}$ to generate the representation $\tilde{\vz_t}$ from the visual observation $\vo_t$ and implement the policy learner to control the agent.

\begin{algorithm}[tb]
	\caption{Training Bidirectional Transition (BiT) Model} 
	\label{alg1} 
	\begin{algorithmic}[1]
		\STATE $\theta,\varphi$: initialize feature extractor parameters
        \STATE $\psi,\upsilon, \nu$: initialize transition learner parameters
        \STATE $\omega $: RL updates per iteration
        \STATE $\epsilon$: momentum coefficient
        \FOR{every iteration}    
        \FOR{update $ = 1, 2, ..., \omega$ }    
            \STATE Sample batch of transitions $(\vo_{t},\va_{t},r_{t},\vo_{t+1}) \sim \mathcal{B}$
            \STATE Obtain features $\tilde{\vz}_{t},\tilde{\vz}_{t+1}$ by $f_{\varphi}$
            \STATE Minimize $\mathcal{L}_{\mathrm{RL}}$ 
        \ENDFOR
        \STATE Sample batch of observations $\vo_{t},\vo_{t+1} \sim \mathcal{B}$
        \STATE Augment observations $\vo^{\prime}_t=Aug(\vo_t) $
        \STATE Obtain online feature $\vz_{t}'$ by $f_{\varphi},g_{\varphi},q_{\varphi}$
        \STATE Obtain target feature $\vz_{t},\vz_{t+1}$ by $f_{\theta},g_{\theta}$
        \STATE Action prediction $\hat{\va} = h_{\psi}(\vz'_{t},\vz_{t+1})$
        \STATE Forward transition prediction $\hat{\vz}_{t+1} = F_{\upsilon}(\vz'_{t}, \hat{\va})$
        \STATE Backward transition prediction $\hat{\vz}_{t} = B_{\nu}(\hat{\va} , \vz_{t+1})$
        \STATE Minimize $\mathcal{L}_{\mathrm{BiT}}$ 
        \STATE Update $\theta \gets (1-\epsilon )\theta + \epsilon\varphi $
        \ENDFOR
	\end{algorithmic} 
\end{algorithm}

\subsection{Feature Extractor}
Given the input observation $\vo \in \mathbb{R}^{\mathrm {C} \times \mathrm {H} \times \mathrm {W}}$, we employ the feature extractor to acquire an encoded representation $\tilde{\vz}$.
Our feature extractor comprises both an online branch and a target branch. 
The online branch encompasses three components, \textit{i.e.,} a shared encoder $f_{\varphi}$, a projection $g_{\varphi}$, and a prediction head $q_{\varphi}$. 
Meanwhile, the target branch includes an encoder $f_{\theta}$ and a projection $g_{\theta}$, sharing the same structure with the corresponding ones of the online branch.

Specifically, we sample the image observations  $\vo_{t}$ and $\vo_{t+1}$ from the buffer $\mathcal{B}$ for every training step. 
We utilize data augmentation to perturb $\vo_{t}$ and produce an augmented observation $\vo'_{t}$.
Subsequently, we derive the corresponding representation $\vz'_{t}=q_{\varphi}(g_{\varphi}(f_{\varphi}(\vo'_{t})))$ through encoding, projection, and prediction.
Identically, we obtain the corresponding representations $\vz_{t}$ and $\vz_{t+1}$ using the target branch: $\vz_{t} = g_{\theta}(f_{\theta}(\vo_{t}))$ and $\vz_{t+1} = g_{\theta}(f_{\theta}(\vo_{t+1}))$.
During training, the encoder $f_{\varphi}$, projection $g_{\varphi}$, and prediction $q_{\varphi}$ are trained simultaneously by the bidirectional transition learner's total objective function $\mathcal{L}_{\mathrm{BiT}}$.
$f_{\varphi}$ is also optimized by the policy learner's objective function $\mathcal{L}_{\mathrm{RL}}$.
For $\theta$, we perform a stop-gradient operation to prevent parameters update through gradient backpropagation.
Specifically, we define the parameters $\theta$ as an Exponential Moving Average (EMA) of $\varphi$, following the rule:
\begin{equation}
\theta_{n+1} \gets (1-\epsilon )\theta_{n} + \epsilon\varphi_{n},
\end{equation}
where $\epsilon \in (0,1]$ is a momentum coefficient, and $n$ is the iteration step of the training process.
This architecture enhances expressiveness in the acquired representation, aligning with prior studies~\cite{chen2020simple,grill2020bootstrap}.

\subsection{Bidirectional Transition Learner}
To obtain a reliable state representation of the current observation, we believe that the representation should be able to forward predict future developments and backward trace histories.
We propose a bidirectional transition learner, where representations generated by the feature extractor are utilized to predict the environment's forward and backward transitions.
This learner consists of three modules:  a forward transition module $F_{\upsilon}$, a backward transition module  $B_{\nu}$, and an action prediction module $h_{\psi}$.

Specifically, we first establish an action prediction module $h_{\psi}$ to derive a pseudo action $\hat{\va}$ based on the input state representations. 
The pseudo action is obtained as $\hat{\va} = h_{\psi}(\vz',\vz_{t+1})$, and the corresponding  optimization objective function is as follows:
\begin{equation}
\mathcal{L}_{\psi }^{\mathrm{action}}=\mathbb{E}_{\tau \sim \mathcal{D}_{\mathrm{env}}}[\left \|\hat{\va} -\va \right \| _{2}^{2}],
\end{equation}
where $\mathcal{D}_{\mathrm{env}}$ is the trajectory dataset of the agent, $\tau$ represents a transition $(\vo_{t},\va_{t},r_{t},\vo_{t+1})$ sampled from $\mathcal{D}_{\mathrm{env}}$, and $\left \|\cdot \right \|_{2}^{2}$ is the mean squared error.
The action prediction module assists the forward and backward transitions without falling into the trivial solution and enhances the task-related information of the representations.

After predicting the pseudo action $\hat{\va}$, we feed it into the forward and backward transition modules. 
Considering the influence of sparse rewards on the representation learning process, our transition model exclusively focuses on predicting state representation transitions without considering reward function.
Specifically, the forward transition module $F_{\upsilon}$ utilizes the augmented representation $\vz'$ and the pseudo action $\hat{\va}$ as inputs, outputting the predicted transition state representation $\hat{\vz}_{t+1}$ at the next time.
The process is $\hat{\vz}_{t+1} = F_{\upsilon}(\vz'_{t}, \hat{\va})$, and the corresponding objective function is defined as follows:
\begin{equation}
\mathcal{L}_{\upsilon }^{\mathrm{fwd}}=\mathbb{E}_{\tau \sim \mathcal{D}_{\mathrm{env}}}[\left \|\hat{\vz}_{t+1}-\vz_{t+1}  \right \|_{2}^{2} ].
\end{equation}
Conversely, the backward transition module $B_{\nu}$ utilizes the state representation $\vz_{t+1}$ at the next step and pseudo action $\hat{a}$ to output the current state representation prediction $\hat{\vz}_{t}$.
The process is expressed as $\hat{\vz}_{t} = B_{\nu}(\hat{\va} , \vz_{t+1})$, and the objective function is defined as follows:
\begin{equation}
\mathcal{L}_{\nu }^{\mathrm{bwd}}=\mathbb{E}_{\tau \sim \mathcal{D}_{\mathrm{env}}}[\left \|\hat{\vz}_{t}-\vz_{t}  \right \|_{2}^{2} ].
\end{equation}

Note that we do not predict environment transition in the pixel space but in the compressed state representation space.
This approach enables the representation to circumvent the uncertainty of predicting pixels and prevents interference from task-independent areas within the pixel space.
We jointly optimize the bidirectional transition learner and the above feature extractor with the total objective function $\mathcal{L}_{\mathrm{BiT}}$ as follows:
\begin{equation}
\mathcal{L}_{\mathrm{BiT}} =\mathcal{L}_{\upsilon  }^{\mathrm{fwd}}+\mathcal{L}_{\nu }^{\mathrm{bwd}}+\mathcal{L}_{\psi  }^{\mathrm{action}}.
\end{equation}
After optimizing the total objective function $\mathcal{L}_{\mathrm{BiT}}$, the extracted representation will more precisely reflect the state information of the agent, providing more reliable support for correct decision-making.

\subsection{Policy Learner}
BiT is independent of the policy learner as an encoding representation module and applies to any visual RL algorithm. 
This paper implements a standard off-policy RL method Soft Actor-Critic (SAC) \cite{haarnoja2018soft} as the policy learner.
Specifically, the unaugmented raw image observation $\vo_t$ is fed into the encoder $f_{\varphi}$ to output the representation $\tilde{\vz}_{t}$. Then, the policy learner utilizes $\tilde{\vz}_{t}$ to train the policy. 
As shown in Algorithm~\ref{alg1}, the optimization of representation learning $\mathcal{L}_{\mathrm{BiT}}$ is performed alternately with the optimization of reinforcement learning $\mathcal{L}_{\mathrm{RL}}$ with $\omega$ steps.
The two optimization objectives jointly update the parameters of the shared encoder $f_{\varphi}$.
We present a more detailed architecture of SAC in the supplementary material.

\begin{table*}
\begin{center}
\resizebox{1\textwidth}{!}{
\begin{tabular}{c|c|ccccccc}
\hline
Setting & Task &CURL&PAD &SODA &SVEA &TLDA &BiT    \\[1pt]\hline
\multirow{6}{*}{Video Easy}    
&Walker Stand &852±75 &935±20  &955±13  &961±8 &\textbf{973±6} &\underline{966±3}  \\[1pt]
&Cartpole Swingup&404±67&521±76&758±62&\textbf{782±27}&671±57&\underline{779±34}  \\[1pt]
&Ball\_in\_cup Catch&316±119&436±55&\underline{875±56}&871±106&855±56&\textbf{899±23}\\[1pt]
&Finger Spin &502±19 &691±80    &695±97  &\underline{808±33} &756±87 &\textbf{835±25}  \\[1pt]
&Reacher Easy &724±49 &692±14    &567±64  &294±43 &\underline{976±1} &\textbf{978±4} \\[1pt]\hline
\multirow{6}{*}{Video Hard}  
&Walker Stand &45±5 &278±72    &771±83  &\textbf{834±76} &769±59 &\underline{829±24} \\[1pt]
&Cartpole Swingup &114±15 &123±24  &\underline{429±64} &393±45 &223±62 &\textbf{526±40}\\[1pt]
&Ball\_in\_cup Catch &115±33 &66±61   &327±100  &403±174 &\underline{457±16} &\textbf{570±71}  \\[1pt]
&Finger Spin &27±21 &56±18 &302±41  &\underline{335±58} &224±56 &\textbf{400±13}  \\[1pt]
&Reacher Easy &86±11 &118±35  &233±57  &\underline{236±56} &160±67 & \textbf{590±43} \\[1pt]\hline
\end{tabular}}
\caption{\textbf{Generalization Performance.}
We report the average episode return and corresponding variances over three different seeds.
These methods are trained on a distractor-free background and tested on different video distractor backgrounds.
Our algorithm attains state-of-the-art performance in most tasks while also delivering competitive results in others.
}
\label{DMCtronl}
\end{center}
\end{table*}

\section{Experiments}
We validate the effectiveness of BiT with other state-of-the-art methods on a wide spectrum of visual generalization tasks.
We mainly choose two configurations of the DeepMind Control suite (DMControl) \cite{tassa2018deepmind,hansen2021generalization} to compare the generalization performance and sample efficiency. 
We also perform ablation and augmentation methods analysis to study BiT.
Moreover, we implement the robotic manipulation simulator \cite{jangir2022look} and driving simulator CARLA  \cite{dosovitskiy2017carla}. 
More experiments details and results on CARLA are in the supplementary material.

\subsection{DMControl Environment }
DMControl is a vision-based simulator offering a collection of continuous control tasks. 
The generalization problem in DMControl introduces visually distracting environments unrelated to reward signals for motor control tasks.
To evaluate the effectiveness of our method, we conduct experiments utilizing two distinct settings to assess generalization capacity and sample efficiency.

\begin{figure}
  \centering
  \includegraphics[width=1\linewidth]{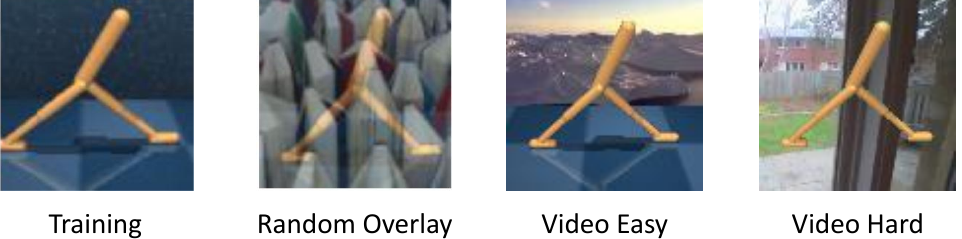}
  \caption{\textbf{Examples of Generalization Setup. } We utilize three different background types,  \textit{i.e.,} Distractor-free for training, ``Video Easy'' and ``Video Hard''. 
  The data augmentation in this environment is ``random overlay''. 
  There are examples of the ``Walker Stand'' task.
  }
  \label{DMC_Base_case}
\end{figure}

\textbf{Generalization Setup. } 
The training environment is without background distractions, and the test environment has two video background interference settings of the DMControl Generalization Benchmark (DMControl-GB) \cite{hansen2021generalization}.
The degree of video impact in the test environment can be divided into ``Video Easy'' and ``Video Hard''.
We verify the generalization of our method on five tasks.
We mainly consider the random overlay to augment the image observation, which is also applied in the prior methods.
Samples from different environments are shown in Figure \ref{DMC_Base_case}.
The comparison methods are as follows: 
(1) CURL \cite{laskin2020curl} implements contrastive learning and data augmentations; 
(2) PAD \cite{DBLP:conf/iclr/HansenJSAAEPW21} implies an auxiliary task for continuing training in a novel environment; 
(3) SODA \cite{hansen2021generalization} utilizes the BYOL-like \cite{grill2020bootstrap} architecture to learn the latent representation consistency; 
(4) SVEA \cite{hansen2021stabilizing} proposes a technique for stabilizing high-variance Q-targets. 
(5) TLDA \cite{DBLP:conf/ijcai/YuanMMX00LX22} reduces the interference of augmentation to the image.
We run 3 random seeds and report the mean and standard deviation of the episode return in Table \ref{DMCtronl}.


\begin{figure}
  \centering
  \includegraphics[width=1\linewidth]{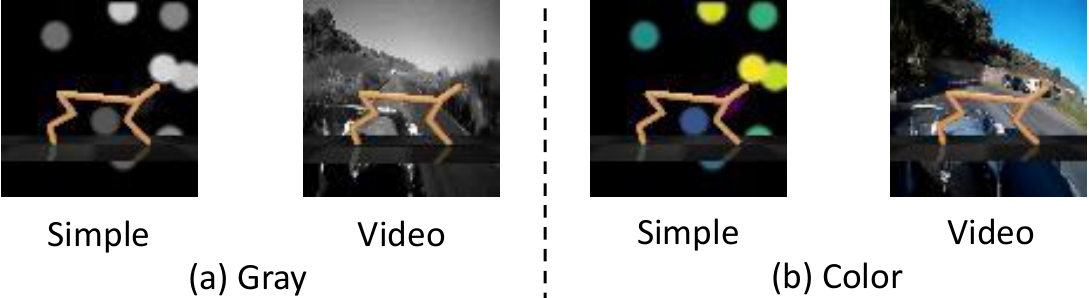}
  \caption{\textbf{Examples of Sample Efficiency Setup.} We choose two color configurations, namely ``Gray'' and ``Color'', for two video backgrounds: ``Simple Distractor'' and ``Natural Video''.
  There are examples of  the ``Cheetah Run'' task.}
  \label{DMC_DBC_case}
\end{figure}

\begin{table*}
\begin{center}
\resizebox{0.95\textwidth}{!}{
\begin{tabular}{c|ccccccc|cc}
\hline
\multicolumn{8}{c}{Gray}\vline & \multicolumn{2}{c}{Color}\\[1pt]\hline
Task & SAC &DrQ &CURL &SODA &PAD &SPD &BiT & SPD & BiT    \\[1pt]\hline
\multicolumn{10}{l}{ (a) Sample Efficiency (training and testing on Simple Distractor)}\\[1pt]\hline
Cheetah Run &230±20 &272±31 &335±1 &304±24 &301±33 &334±3 &\textbf{349±26} &317±25 &\textbf{333±42}  \\[1pt]
Finger Spin &399±25 &665±27 &656±47 &735±33 &689±27 &\textbf{984±1} &983±1 &\textbf{981±4} &977±11  \\[1pt]
Hopper Hop  &92±6 &92±32 &74±28 &86±44 &125±87 &153±6 &\textbf{234±20} &\textbf{181±74} &138±7 \\[1pt]
Reacher Easy&107±1 &230±47 &409±45 &286±50 &287±160 &646±107 &\textbf{953±36} &590±170 &\textbf{778±101} \\[1pt]
Walker Walk &37±5 &494±105 &917±12 &869±12 &862±2 &895±7 &\textbf{917±59} &927±18 &\textbf{928±24}  \\[1pt]\hline
\multicolumn{10}{l}{(b) Sample Efficiency (training and testing on Natural Video)}\\[1pt]\hline
Cheetah Run &136±22 &64±20 &118±38 &74±31 &171±114 &330±26 &\textbf{411±13} &81±47 &\textbf{387±200}  \\[1pt]
Finger Spin &289±12 &205±145 &227±147 &59±40 &3±2 &\textbf{983±1} &712±2 &171±25 &\textbf{728±180}  \\[1pt]
Hopper Hop  &33±7 &0±0 &10±5 &0±0 &1±1 &\textbf{164±14} &41±15 &18±15 &\textbf{70±37} \\[1pt]
Reacher Easy&100±2 &90±15 &414±107 &80±5 &104±21 &574±62 &\textbf{633±73} &190±152 &\textbf{605±237} \\[1pt]
Walker Walk &33±2 &104±43 &812±52 &404±47 &73±8 &896±18 &\textbf{947±5} &571±163 &\textbf{654±173}  \\[1pt]\hline
\multicolumn{10}{l}{(c) Generalization (training on Simple Distractor but testing on Natural Video)}\\[1pt]\hline
Cheetah Run &51±18 &219±25 &190±31 &229±18 &298±19 &329±6 &\textbf{331±36} &174±39 &\textbf{306±32}  \\[1pt]
Finger Spin&125±27 &661±27 &647±44 &653±39 &690±28 &893±30 &\textbf{974±19} &113±75 &\textbf{983±2}  \\[1pt]
Hopper Hop&15±5 &81±30 &43±24 &59±32 &113±68 &135±6 &\textbf{197±18} &22±2 &\textbf{138±1} \\[1pt]
Reacher Easy&109±4 &158±20 &287±47 &160±28 &274±158 &432±119 &\textbf{844±19} &170±43 &\textbf{860±31} \\[1pt]
Walker Walk &58±17 &271±82 &408±35 &754±25 &835±1 &855±16 &\textbf{927±4} &634±34 &\textbf{876±31}  \\[1pt]\hline
\end{tabular}}
\caption{\textbf{Sample Efficiency and Generalization Performance.}
We report the average episode return and corresponding variances for over three seeds.
We train methods on ``Simple Distractor'' and ``Natural Video''. 
We evaluate methods on the same background for sample efficiency experiments and also assess methods training on Simple Distractor on novel Natural Video for generalization experiments.
Our method achieves state-of-the-art results on most tasks, particularly in the more challenging setting, ``Color Natural Video''.
}
\label{DMsuiteSPD}
\end{center}
\end{table*}

\textbf{Sample Efficiency Setup. }    
We train our method using four interference settings and five tasks, following the approach outlined in \cite{zhang2020learning}, to assess the anti-interference capability and sample efficiency during training.
Specifically, we employ two distinct video backgrounds: ``Simple Distractor'' and ``Natural Video'', along with two color settings: ``Gray'' and ``Color''.
Examples from different settings are shown in Figure \ref{DMC_DBC_case}.
The testing and training environment background settings are the same for the sample efficiency assessment.
Furthermore, we conduct a generalization evaluation in this setup.
We train methods in the simple distractor background and compare their performance in the novel natural video background.

For a fair comparison of the method's inherent generalization ability rather than the impact of the data augmentation technique, we implement the same data augmentation method as SPD.
We apply weak and strong augmentation methods to the image observations, respectively.
In this section, we compare our approach with CURL, SODA, PAD, and other new methods as follows:
SAC \cite{haarnoja2018soft} is a standard off-policy RL algorithm.
DrQ \cite{yarats2020image} incorporates multiple image transformations.
SPD \cite{kim2022self} employs a self-predictive forward dynamics approach to efficiently extract task-relevant features.
The comparative results are carried out utilizing 3 random seeds and reported in Table \ref{DMsuiteSPD}.

\textbf{Generalization Performance. } 
The results in distraction-free background training and evaluation with video distractors in five DMControl-GB tasks are presented in Table \ref{DMCtronl}. 
Our method excels in most tasks, achieving optimal values in 7 out of 10 tasks and suboptimal values in the remaining 3 tasks. 
In contrast, alternative algorithms, such as SODA, SVEA, or TLDA, only exhibit superior performance in a few specific tasks, lacking the capability to achieve optimal results across the majority of tasks.
The representations learned by our method enable a correct and reliable comprehension of environmental observations and effectively generalize to novel environments.
As evidenced in Table \ref{DMsuiteSPD} (c), even when trained in the presence of noisy backgrounds, BiT retains the capacity to acquire a dependable comprehension of the environment, showcasing robust generalization skills. 
Our model consistently outperforms state-of-the-art algorithms across all tasks that were trained on the simple distractor background and then generalized to the natural video background.
In contrast, other algorithms frequently struggle to learn reliable representations and effective policies when confronted with interfering backgrounds.

\textbf{Sample efficiency Performance. } 
As presented in Table \ref{DMCtronl} (a) and (b), the training and test environments are identical, and we compare the training sample efficiency.
Our method surpasses other baselines on 7 out of 10 tasks in the context of the ``Simple Distractor'' background. 
Furthermore, BiT outperforms other baselines on 8 out of 10 tasks when considering the ``Natural Video'' background. 
Notably, BiT attains state-of-the-art performance on the particularly challenging task: ``Color'' and ``Natural Video'' background distraction.
Specifically analyzing SPD, its generalization ability significantly diminishes when changing the training environment from gray to color. 
This disparity can be attributed to the fact that SPD employs forward transition for representation learning, thereby hindered by the uncertainties inherent to unidirectional transfer predictions. 
Consequently, SPD fails to furnish stable representations conducive to policy learning.
In contrast, our bidirectional transition learner simultaneously encompasses both forward and backward transition modules as auxiliary tasks. 
This bidirectional approach substantially mitigates the instability of representation learning and provides a dependable state representation, thus enhancing the efficacy of agent policy learning. 
As a result, our model demonstrates superior sample efficiency and enhanced generalization capability.
In the supplementary material, we further compare the learning curve with the SPD algorithm to analyze the sample efficiency.

\begin{figure}
  \centering
  \includegraphics[width=1\linewidth]{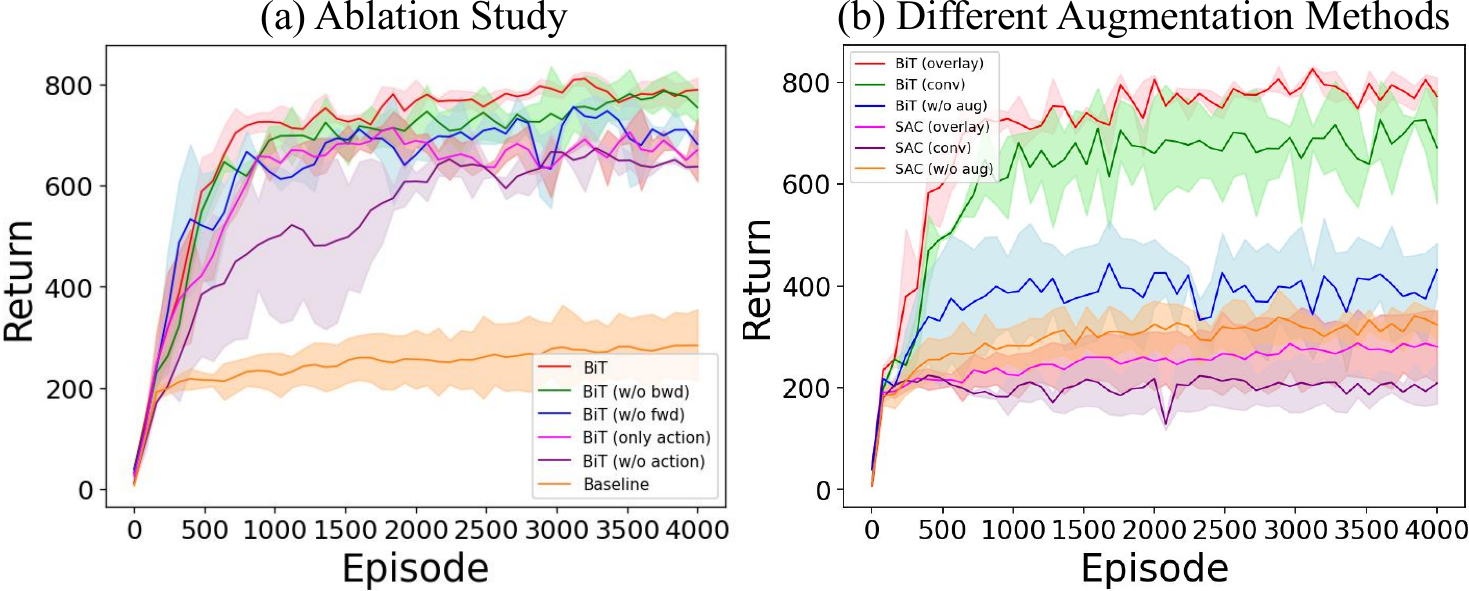}
  \caption{\textbf{Left: Ablation Study.} In Figure (a), we analyze the effectiveness of each constraint objective.
  \textbf{Right: Different Augmentation Methods.} In Figure  (b), we assess the impact of data augmentation.
  }
  \label{Ablation}
\end{figure}

\textbf{Ablation Study. }
We conduct ablation experiments on the DMControl-GB task to verify the roles of each objective function in our bidirectional transition learner, as shown in Figure \ref{Ablation} (a).
We chose the ``Cartpole Swingup'' task under the ``Video Easy'' background.
``BiT w/o fwd'' indicates the absence of a forward transition module, while ``BiT w/o bwd'' signifies the omission of a backward transition module. 
``BiT w/o action'' implies the exclusion of an action prediction module, and ``BiT only action'' suggests the presence of only the action prediction module within the bidirectional transition learner. 
The term ``Baseline'' refers to a degraded version of our model, lacking the proposed objective function but including data augmentation.

Figure \ref{Ablation} shows that BiT yields the best generalization effect and sample efficiency.
We found that adding forward or backward transition prediction separately on action prediction also improves performance, but the enhancement is insignificant.
That is because of the unidirectional transition prediction's inherent instability, and the agent might lead to a biased comprehension of the observation.
In contrast, bidirectional prediction helps the agent with a more stable state representation and mitigates the uncertainty of observational comprehension by concurrently predicting both forward and backward transition. 
This principle aligns with the prior works \cite{lai2020bidirectional, lyu2022double}.
Moreover, omitting the action prediction module cause the learned representation to lack task-related information and easily fall into a trivial solution.
Overall, the result demonstrates that each module in BiT plays a significant and effective role.

\textbf{Study of Different Augmentation Methods. } 
We analyze the effects of different data augmentations for the model, \textit{i.e.,} ``random overlay,'' ``random convolution,'' and ``without augmentation''.
We conduct a comparison with the base SAC algorithm.
The task setting is the same as the ablation study section, and the results are depicted in Figure \ref{Ablation} (b).

\begin{figure}
  \centering
  \includegraphics[width=0.9\linewidth]{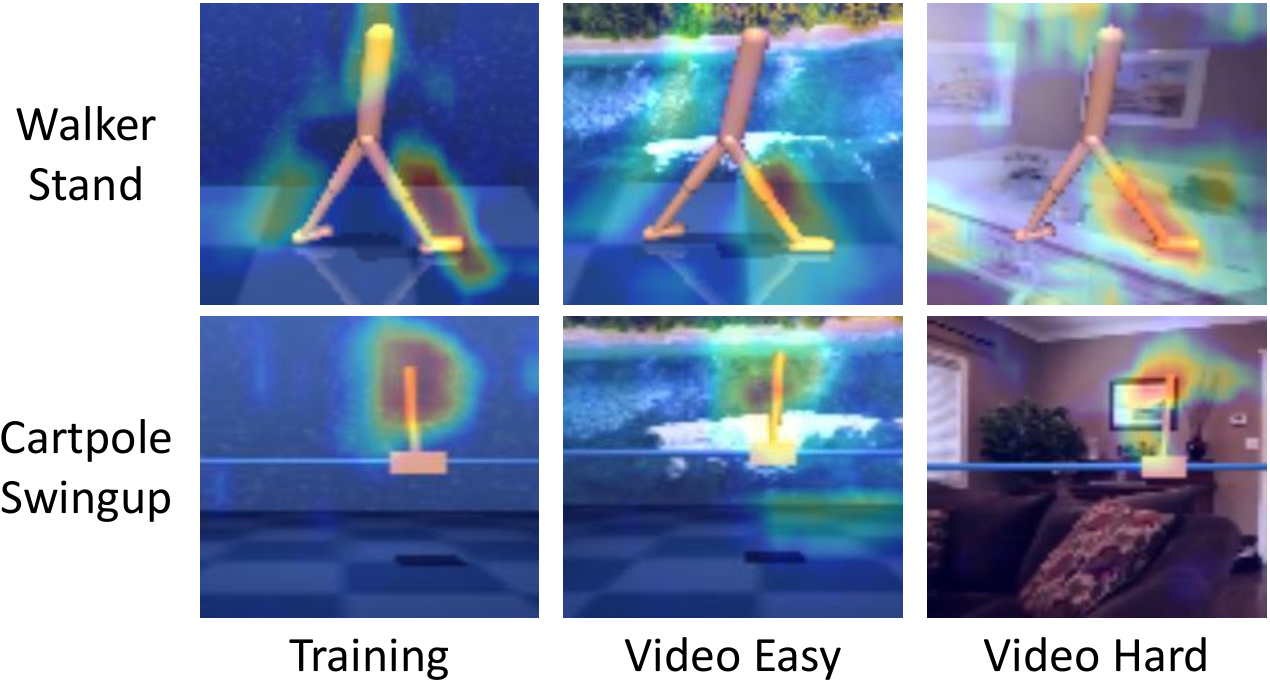}
  \caption{\textbf{Attention Maps of BiT.} We show the attention maps of BiT on the different backgrounds of tasks ``Walker walk'' and ``Cartpole Swingup''. 
  }
  \label{attention_map}
\end{figure}
The results demonstrate that our model achieves substantial improvements over the SAC, both with and without data augmentation.
With data augmentation, the SAC algorithm struggles to acquire a stable representation due to the increased diversity in the samples.
Consequently, its ability to improve generalization is limited, and in some cases, it even results in performance degradation.
In contrast, BiT effectively acquires a stable and generalizable representation through data augmentation.
Moreover, BiT has a larger performance gain utilizing the ``random overlay''.
It is because the ``random overlay'' produces texture increases that better match the video background than the color distractions induced by a ``random convolution''.

\textbf{Visualization of Attention Map.} 
We choose two tasks in the generalization setting: ``Walker Stand'' and ``Cartpole Swingup'', to show the attention maps of BiT on different backgrounds, \textit{i.e.,} ``distracting-free'', ``Video Easy'' and ``Video Hard''.
The results in Figure \ref{attention_map} show that BiT can clearly perceive the task-related regions in the image observation, regardless of the presence or absence of background interference.
In this way, BiT provides a reliable and correct representation of the image observation for the agent to learn a generalizable policy.

\subsection{Robotic Manipulation Environment}
To demonstrate the wide applicability of our method, we implement two tasks from the vision-based robotic manipulation simulator \cite{jangir2022look}.
We train agents on the default background and validate them on five testing environments.
The test environments are similar to the training ones but with different colors and textures for the background. 
The evaluated algorithms include SAC, SODA, and SVEA, with the corresponding results  presented in Table \ref{robotsuite}.
Experimental results demonstrate that BiT can well generalize in robotic training domain novel environments.

\begin{table}
\begin{center}
\resizebox{0.5\textwidth}{!}{
\begin{tabular}{c|c|cccc}
\hline
Method & Environment&SAC &SODA &SVEA &BiT    \\[1pt]\hline
\multirow{6}{*}{Reach}    
&test\_1 & -29±16 &-28±28 &-39±25 &\textbf{-9±18} \\[1pt]
&test\_2 &-26±21 &-31±18 &-7±36 &\textbf{-3±21} \\[1pt]
&test\_3 &-41±15 &-35±15 &\textbf{-16±36} &-28±30 \\[1pt]
&test\_4 &-23±23 &-37±22 &\textbf{-5±32} &-27±34 \\[1pt]
&test\_5 &-39±11 &-55±14 &-49±23 &\textbf{-25±23} \\[1pt]
&Avg &-32±17 &-37±19 &-23±30 &\textbf{-18±25} \\[1pt] \hline
\multirow{6}{*}{Peg in box}  
&test\_1 &-55±8 &-16±76 &-40±47 &\textbf{-16±57} \\[1pt]
&test\_2 &-62±11 &-10±68 &18±67 &\textbf{27±43} \\[1pt]
&test\_3 &-53±6 &-1±91 &16±33 &\textbf{55±52} \\[1pt]
&test\_4 &-61±11 &-27±81 &48±60 &\textbf{72±48} \\[1pt]
&test\_5 &-66±21 &-87±50 &-51±42 &\textbf{1±59} \\[1pt]
&Avg &-59±11 &-28±73 &-2±50 &\textbf{28±52} \\[1pt]\hline 
\end{tabular}}
\caption{\textbf{Performance on the robotic environments.}
We report the average return and variance for each task and their mean.
Our algorithm excels in most tasks, demonstrating the wide applicability of our model.
}
\label{robotsuite}
\end{center}
\end{table}


\section{Conclusion}
In this paper, we introduce a \textbf{Bi}directional \textbf{T}ransition (BiT) Model to aid agents in extracting reliable and well-generalized representations from visual observations.
The BiT method utilizes the bidirectional transition prediction objectives to ensure the agent accurately comprehends visual observations.
Our method demonstrates competitive generalization and sample efficiency on various benchmark tasks,  including the DeepMind Control suite, robotic manipulation simulators, and driving simulator CARLA.

\bibliography{aaai24}

\end{document}